\documentclass[10pt,journal,compsoc]{IEEEtran}

\ifCLASSOPTIONcompsoc
  \usepackage[nocompress]{cite}
\else
  \usepackage{cite}
\fi
\ifCLASSINFOpdf
\else
\fi
\usepackage{cite}
\usepackage{multirow}
\usepackage{graphicx}
\usepackage{subfigure}
\usepackage{color}
\usepackage{dcolumn}
\usepackage{amsmath}
\usepackage{algorithm}
\usepackage{algorithmic}
\usepackage{verbatim}
\usepackage{float}
\begin{document}

\title{Incremental Learning Using a Grow-and-Prune Paradigm with Efficient Neural Networks}

\author{Xiaoliang~Dai,~Hongxu~Yin,~and~Niraj~K.~Jha,~\IEEEmembership{Fellow,~IEEE}
\thanks{This work was supported by NSF Grant No. CNS-1617640. Xiaoliang Dai, 
Hongxu Yin, and Niraj K. Jha are with the Department
of Electrical Engineering, Princeton University, Princeton,
NJ, 08544 USA, e-mail:\{xdai, hongxuy, jha\}@princeton.edu.}}

\IEEEtitleabstractindextext{%
\begin{abstract}
Deep neural networks (DNNs) have become a widely deployed model for numerous machine learning 
applications.  However, their fixed architecture, substantial training cost, and significant model 
redundancy make it difficult to efficiently update them to accommodate previously unseen data.  To 
solve these problems, we propose an incremental learning framework based on a grow-and-prune 
neural network synthesis paradigm.  When new data arrive, the neural network first grows new 
connections based on the gradients to increase the network capacity to accommodate new data.  Then, 
the framework iteratively prunes away connections based on the magnitude of weights to 
enhance network compactness, and hence recover efficiency.  Finally, the model rests at a lightweight 
DNN that is both ready for inference and suitable for future grow-and-prune updates.  The proposed 
framework improves accuracy, shrinks network size, and significantly reduces the additional training 
cost for incoming data compared to conventional approaches, such as training from scratch and network 
fine-tuning.  For the LeNet-300-100 and LeNet-5 neural network architectures derived for the MNIST 
dataset, the framework reduces training cost by up to 64\% (63\%) and 67\% (63\%) compared to 
training from scratch (network fine-tuning), respectively.  For the ResNet-18 architecture
derived for the ImageNet dataset and DeepSpeech2 for the AN4 dataset, the corresponding training 
cost reductions against training from scratch (network fine-tunning) are 64\% (60\%) and 67\% (62\%), 
respectively. Our derived models contain fewer network parameters but achieve higher accuracy 
relative to conventional baselines.
\end{abstract}

}

\maketitle

\IEEEdisplaynontitleabstractindextext

\IEEEpeerreviewmaketitle

\IEEEraisesectionheading{\section{Introduction}\label{sec:introduction}}
In recent years, deep neural networks (DNNs) have achieved remarkable success and emerged as an 
extraordinarily powerful tool for a wide range of machine learning applications.  Their ability 
to represent input data through increasingly more abstract layers of feature representations and 
knowledge distillation has been shown to be extremely effective in numerous application areas, such 
as image recognition, speech recognition, disease diagnosis, and neural machine 
translation~\cite{AlexNet, deepspeech2, fMRI, fMRI2, translation, voice}.  With increased access to 
large amounts of labeled training data (e.g., ImageNet~\cite{imageNetdataset} with 1.2 million 
training images from 1,000 different categories) and computational resources, DNNs have
resulted in human-like or even super-human performance on a variety of tasks.

A typical development process of a DNN starts with training a model based on the target dataset 
that contains a large amount of labeled training instances.  The DNN learns to distill intelligence 
and extract features from the dataset in this process.  The well-trained model is then used to make 
predictions for incoming unseen data~\cite{incremental0}.  In such a setting, all the labeled data 
are presented to the network all-at-once for one training session. While effective, this may be too 
idealized for many real-world scenarios where training data and their associated labels may be 
collected in a continuous and incremental manner, and only some data instances may be used initially 
to obtain the first trained model. For example, biomedical datasets are typically updated regularly 
when the number of data points obtained from patients increases, or disease trends shift across 
time~\cite{hdss}. This makes it necessary to update a DNN model frequently to accommodate the new 
data and capture the new information effectively.

A widely-used approach for updating DNNs to learn new information involves discarding the existing 
model and retraining the DNN weights from scratch using all the data acquired so 
far~\cite{learn_plus, incremental0}.  This method leads to a complete loss of all the previously 
accumulated knowledge in the pre-trained network, and suffers from three major problems:

\begin{itemize}
\item \textbf{Vast training cost}: Training from scratch at each update is computationally- 
and time-intensive. Ideally, an incremental learning system should combine existing knowledge with 
new knowledge in a continuous and efficient manner, hence minimizing additional computational 
costs of an update.
\item \textbf{Fixed network capacity}: Conventional DNN models have fixed and static 
architectures.  As new data become available, it is not possible to increase their capacity during 
the entire training process.  
\item \textbf{Massive redundancy}: The generated DNN models derived for large real-world datasets 
are typically over-parameterized~\cite{PruningHS} and can easily contain millions of parameters, as 
shown in Table~\ref{tabs:params}. Such a large model size incurs substantial storage and memory 
cost during inference.
\end{itemize}

\begin{table}[t]
\caption{Model size of DNNs for real-world applications}
\label{tabs:params}
\begin{center}
\begin{tabular}{lcccc}
\hline
Network & Task & \#Parameters \\
\hline
AlexNet~\cite{AlexNet} & Image classification & 61M &  \\
VGG-16~\cite{VGG} & Image classification & 138M\\
ResNet-18\cite{ResNet} & Image classification & 12M\\
ResNet-152~\cite{ResNet} & Image classification & 60M \\
DeepSpeech2~\cite{stanford} & Speech recognition & 50M \\
Seq2Seq~\cite{translation} & Machine translation & 384M \\
\hline
\end{tabular}
\end{center}
\vskip -0.1in
\end{table}

\begin{figure*}
\begin {center}
\includegraphics[width=168mm]{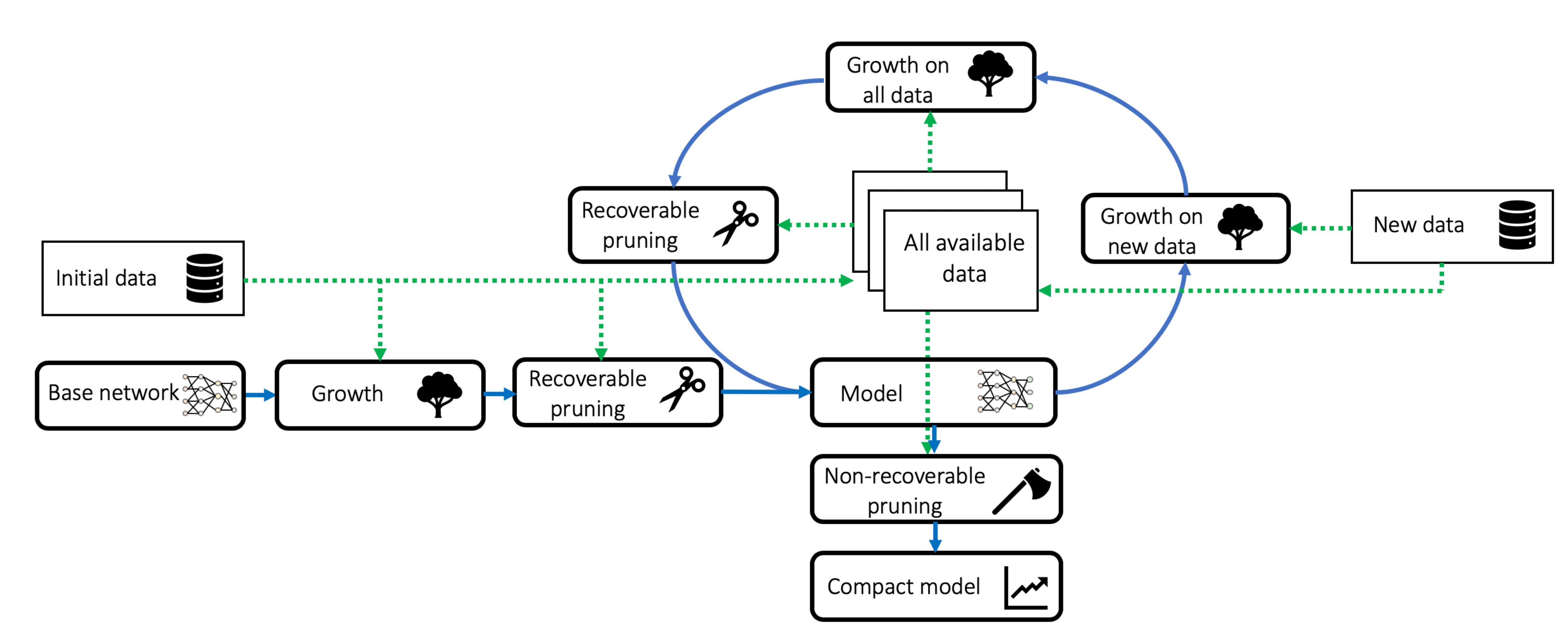}
\end{center}
\caption{Flowchart of the proposed incremental framework.}
\label{fig:flowchart}
\end{figure*}

To address the above problems, we propose an incremental learning framework based on a grow-and-prune 
neural network synthesis paradigm.  It consists of two sequential training stages in a model update 
process: gradient-based growth and magnitude-based pruning.  We depict the flowchart of the 
framework in Fig.~\ref{fig:flowchart}, where the green dashed and blue solid lines depict the data 
flow path and model update process, respectively.  We first grow and prune a model with the initial 
data.  When new data arrive, the network undergoes a growth phase (first, based on new data and then 
on all available data) that increases its size to accommodate new data and knowledge.  Then, we 
employ a pruning phase to remove redundant parameters to obtain a compact inference model.  In 
this stage, we first use a concept called recoverable pruning to acquire a compact model that
is subjected to the next grow-and-prune update, and then use non-recoverable pruning to achieve 
ultra compactness if the use scenario imposes a very strict resource constraint.  We validate our 
approach across different DNN architectures, datasets, and learning tasks, including LeNets on the 
MNIST dataset, ResNet-18 on the ImageNet dataset, and DeepSpeech2 on the AN4 dataset. Our
incremental learning framework consistently leads to improved accuracy, faster model training, and 
more compact final inference models relative to conventional approaches, such as training from 
scratch (TFS) and network fine-tuning (NFT).

The rest of this paper is organized as follows. We review related work in 
Section~\ref{sec:related_work}. In Section~\ref{sec:background}, we introduce background material 
that describes the scope and goal of incremental learning as well as hidden-layer long short-term 
memory (H-LSTM)~\cite{hlstm} cells that we use in our experiments. Then, we discuss our proposed 
incremental learning methodology in detail in Section~\ref{sec:methodology}. 
In Section~\ref{sec:experiments}, we present experimental results for both image classification and 
speech recognition tasks.  In Section~\ref{sec:discussions}, we discuss the inspirations of our 
proposed framework from the human brain.  Finally, we draw conclusions in Section~\ref{sec:conclusion}.

\section{Related work}\label{sec:related_work}
Various methods and algorithms have been proposed in the past to design DNN-based incremental 
learning frameworks and execution-efficient DNNs. We discuss these approaches next.

\subsection {Incremental learning}
The world of digitized data generates new information at each moment, thus fueling the need for 
machine learning models that can learn as the new data arrive.  We summarize such techniques
next.

\noindent \textbf{Transfer learning based approaches}:  Transfer learning provides a promising 
solution to the problem of efficient incremental learning with DNNs.  It effectively conserves 
existing knowledge by maintaining the weights and connections of the first several convolutional 
layers, which are known to be generic feature extractors~\cite{transfer}.  For example, Li et 
al.~propose a framework called `Learning without Forgeting' to train the transferred network with 
only new data while achieving performance improvements.  This also obviates the need to learn from 
scratch for new tasks~\cite{transferlearning1}.  Yan et al.~exploit common feature sharing in 
a hierarchical convolutional neural network (CNN) model to achieve better performance compared 
to their NFT baseline~\cite{hdcnn}.  Another promising approach to efficient incremental learning 
is to transfer knowledge from a small network to a large one.  This yields significant training 
cost reduction and accuracy gain, as shown by Chen et al.~\cite{net2net}.
 
\noindent \textbf{Architectural evolution approaches}: Another way to accommodate new data is to 
adaptively evolve the network architecture and update the weights simultaneously.  Xiao et 
al.~propose an incremental training algorithm that grows a tree-structured network 
hierarchically~\cite{error_driven}.  Roy et al.~also utilize such tree-structured networks to reduce 
the training cost by 20\%~\cite{tree}.  Alternative methods focus on adding new layers or new nodes 
to increase network capacity.  For instance, Rusu et al.~suggest that a progressive neural network 
can learn to solve complex sequences of tasks by adding additional layers and leveraging prior 
knowledge with lateral connections~\cite{progressive_network}.  Terekhov et al.~also propose an 
algorithm to reuse existing network capacity and add new blocks of neurons, where the generated model 
outperforms the baseline network that is trained from scratch~\cite{knowledge}.

\subsection {Efficient and compact neural networks}
Most DNNs are computationally intensive and over-parameterized~\cite{PruningHS}.  Several different 
approaches have been put forward in the literature to design efficient DNNs.  These approaches 
include designing novel compact neural network architectures and compressing existing models.  We 
summarize them next.

\noindent \textbf{Compact architecture design}:
Exploiting efficient building blocks and operations can significantly cut down on the DNN computation 
cost.  For example, MobileNetV2 effectively shrinks the model size and cuts down the number of 
floating-point operations (FLOPs) with inverted residual building blocks~\cite{mobilenetv2}.  Ma 
et al.~propose another compact CNN architecture that utilizes channel shuffle operation and 
depth-wise convolution~\cite{shufflenetv2}.  Wu et al.~suggest replacing spatial convolution layers 
with shift-based modules that have zero FLOPs.  The generated ShiftNet has substantially
reduced computation and storage costs~\cite{shift}. Besides, automated compact architecture design 
also provides a promising solution~\cite{mnasnet}.  Dai et al.~develop an automated architecture 
adaptation and search framework based on efficient performance predictors \cite{chamnet}.  The 
searched models deliver up to 8.5\% absolute top-1 accuracy gain on the ImageNet dataset compared 
to MobileNetV2 while reducing latency.

\noindent \textbf{Model compression}:
Apart from compact architecture design, compressing and simplifying existing models have also emerged 
as a promising approach~\cite{scann}.  By removing redundant connections and neurons, network pruning 
has been shown to be very successful at DNN compression.  For instance, Han et al.~have shown that 
more than 92\% of the connections in VGG-16 can be pruned away without any accuracy 
loss~\cite{PruningHS}.  A recent work combines pruning with network growth and improves the 
compression ratio of VGG-16 by another 2.5$\times$~\cite{nest}.  Furthermore, structured sparsity and 
pruning can reduce the run-time latency significantly~\cite{structuredsparsity}.  For example, 
Yin et al.~achieve $2.4\times$ latency reduction for speech recognition on an Nvidia GPU based on 
hardware-aware structured column and row sparsity~\cite{hardwareprune}.  Besides, low-bit quantization 
is another powerful tool for reducing the storage cost~\cite{deepcompression}.  For instance, Zhu 
et al.~show that replacing a full-precision (32-bit) weight representation with ternary weight 
quantization only incurs a minor accuracy loss for ResNet-18, but significantly reduces the storage 
and memory costs~\cite{tenary}.  

\section{Background}\label{sec:background}
In this section, we discuss background material on the scope and goal of incremental learning and 
H-LSTMs that were used in some of the experiments.

\subsection{Scope and aim of incremental learning}
Incremental learning refers to the process of learning when input data gradually become 
available~\cite{incremental_conv}.  The goal of incremental learning is to let the machine learning 
model preserve existing knowledge and adapt to new data at the same time.  However, aiming to
achieve these two goals simultaneously suffers from the well-known \textit{stability-plasticity 
dilemma}~\cite{dilemma}: a purely stable model is able to conserve all prior knowledge, but cannot 
accommodate any new data or information, whereas a completely plastic model has the opposite problem.

Ideally, an incremental learning framework should have the following 
characteristics~\cite{learn_plus}:
\begin{itemize}
\item \textbf{Flexible capacity:} It should be able to dynamically adjust the model's learning 
capability to accommodate newly available data and information.
\item \textbf{Efficient update:} Updating the model when new data become available should be 
efficient and incur only minimal overhead.
\item \textbf{Preserving knowledge:} It should maintain existing knowledge in the update process, 
and avoid restarting training from scratch.
\end{itemize}

In this work, we introduce another design aim for DNN-based incremental learning systems:
\begin{itemize}
\item \textbf{Compact inference model:} It is beneficial to generate a lightweight DNN model for 
efficient inference.
\end{itemize}
We will show later how our framework addresses the \textit{stability-plasticity dilemma} and 
satisfies all the above requirements.

\subsection{Hidden-layer LSTM}
\begin{figure}
\begin {center}
\includegraphics[width=70mm]{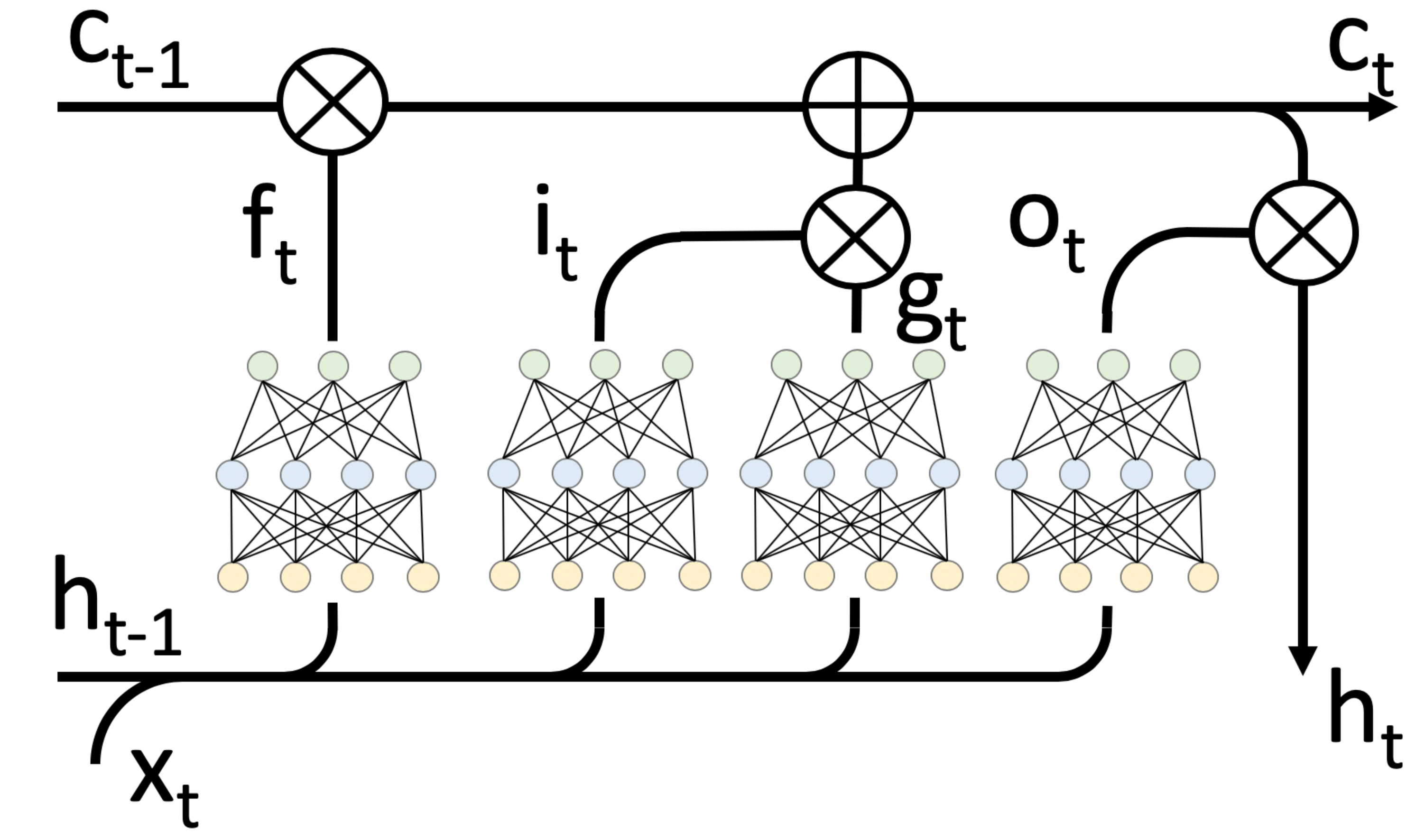}
\end{center}
\caption{Schematic diagram of an H-LSTM cell.  MLP neural networks are used in the control gates.}
\label{fig:hlstm}
\end{figure}

As mentioned earlier, we use the H-LSTM concept proposed in~\cite{hlstm} in our experiments for 
time-series data analysis.  An H-LSTM is an LSTM variant with improved performance and efficiency.  
It introduces multi-level abstraction in the control gates of conventional LSTMs that utilize 
multi-layer perceptron (MLP) neural networks, as shown in Fig.~\ref{fig:hlstm}, where 
$\textbf{c}_{t-1}$ and $\textbf{h}_{t-1}$ refer to the cell state tensor and hidden state
tensor, respectively, at step $t-1$; $\textbf{x}_{t}$, $\textbf{h}_{t}$, $\textbf{c}_{t}$, 
$\textbf{f}_{t}$, $\textbf{i}_{t}$, $\textbf{o}_{t}$, and $\textbf{g}_{t}$, refer to the input 
tensor, hidden state tensor, cell state tensor, forget gate, input gate, output gate, and tensor 
for cell updates at step $t$, respectively; and $\otimes$ and $\oplus$ refer to the
element-wise multiplication operator and element-wise addition operator, respectively.

The MLP gates in an H-LSTM enhance gate control and increase the learning capability of the
cell.  Moreover, they enable drop-out to be used to optimize the control gates and thus alleviate the 
regularization difficulty problem faced by conventional LSTM cells~\cite{hlstm}.  As a result, 
an H-LSTM based recurrent neural network (RNN) achieves higher accuracy with much fewer parameters 
and lower run-time latency compared to an LSTM based RNN for many applications, e.g., image 
captioning and speech recognition.

\section{Methodology}\label{sec:methodology}

In this section, we discuss our proposed incremental learning framework in detail.  We first give 
a high-level overview of the framework, and then detailed descriptions of the specific growth and 
pruning algorithms.

\subsection{Incremental learning framework}

\begin{figure*}
\begin {center}
\includegraphics[width=138mm]{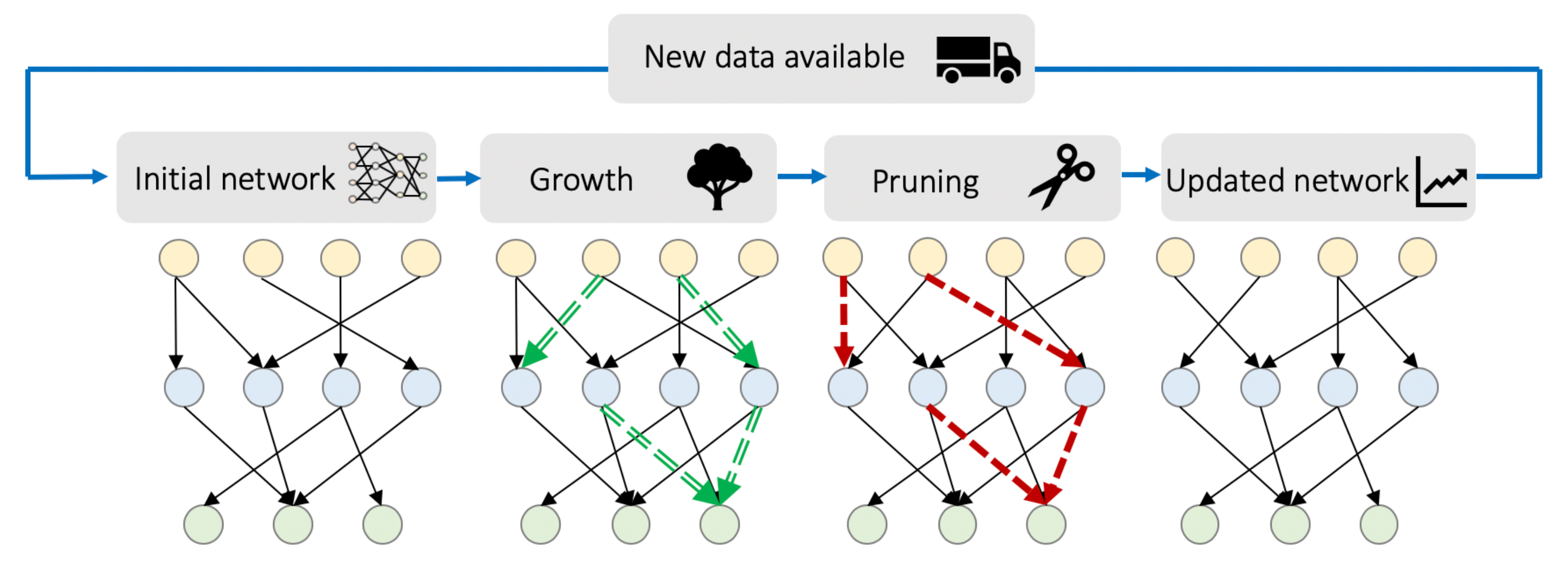}
\end{center}
\caption{Illustration of the grow-and-prune paradigm in the proposed incremental learning framework.}
\label{fig:grow_and_prune}
\end{figure*}

As mentioned earlier, the proposed framework is based on a grow-and-prune paradigm, which enables 
the model to dynamically and adaptively adjust its architecture to accommodate new data and 
information.  We illustrate the growth and pruning process in Fig.~\ref{fig:grow_and_prune}, where 
the double and single dashed lines refer to the newly grown and pruned connections, respectively.  The 
initial network inherits the architecture and weights from the model derived in the last update 
(or uses random weight initialization when starting from scratch for the first model). In the model 
update process, our framework utilizes two sequential steps to update the DNN model: gradient-based 
growth and magnitude-based pruning.  The network gradually grows new connections based on the 
gradient information (extracted using the back-propagation algorithm) obtained in the growth phase.  
Then, it iteratively removes redundant connections based on their magnitudes in the pruning phase.  
Finally, it rests at a compact and accurate inference model that is ready for deployment and the 
next update.  

Next, we explain the gradient-based growth and magnitude-based pruning procedures in detail.

\subsection{Growth phase}
\subsubsection{Gradient-based growth policy}
When new data become available, we use a gradient-base growth approach to adaptively increase the 
network capacity in order to accommodate new knowledge.  The pre-growth network is typically a 
sparse and partially-connected DNN.  In our current implementation, we use a mask tensor 
\textbf{Msk} to disregard the `dangling' connections (connections that are not used in the network) 
for each weight tensor \textbf{W}.  \textbf{Msk} tensors only have binary values (0 or 1) and
have the same size as their corresponding \textbf{W} tensor.  

We employ three sequential steps to grow new connections:
\begin{itemize}
\item \textbf{Gradient evaluation}:  We first evaluate the gradient for all the `dangling' 
connections. In the network training process, we extract the gradient of all weights 
($\textbf{W}.grad$) for each mini-batch of training data with the back-propagation algorithm.  We 
repeat this process and accumulate $\textbf{W}.grad$ over a whole training epoch.  Then, we calculate 
the average gradients over the entire epoch.  Note that we pause the parameter update in the gradient 
evaluation procedure.
\item \textbf{Connection growth}: We activate the connections with large gradients.  Specifically, 
we activate a connection $w$ by manually setting the value of its corresponding mask to be 1 if and 
only if the following condition is met:
\begin{equation}
|w.grad| \geq \alpha ^{th} \text{percentile of} \ \ |\textbf{W}.grad|
\end{equation}
where $\alpha$ is a pre-defined parameter.  We typically use $30 \leq \alpha \leq 50$ in our 
experiments.  This policy helps us activate connections that are the most efficient at reducing the 
loss function $L$.  This is because connections with large gradients also have large derivatives of 
$L$:
\begin{equation}
w.grad = \frac{\partial L}{\partial w}
\end{equation}
\item \textbf{Weight initialization}: We initialize the weights of newly added connections to
$\eta \times w.grad$, where $\eta$ is the current learning rate for training.
\end{itemize}
Connection growth and parameter training are interleaved in the growth phase, where we
periodically conduct connection growth during training.  We employ stochastic gradient descent in both 
the architecture space and parameter space in this process.

\begin{figure}
\begin {center}
\includegraphics[width=66mm]{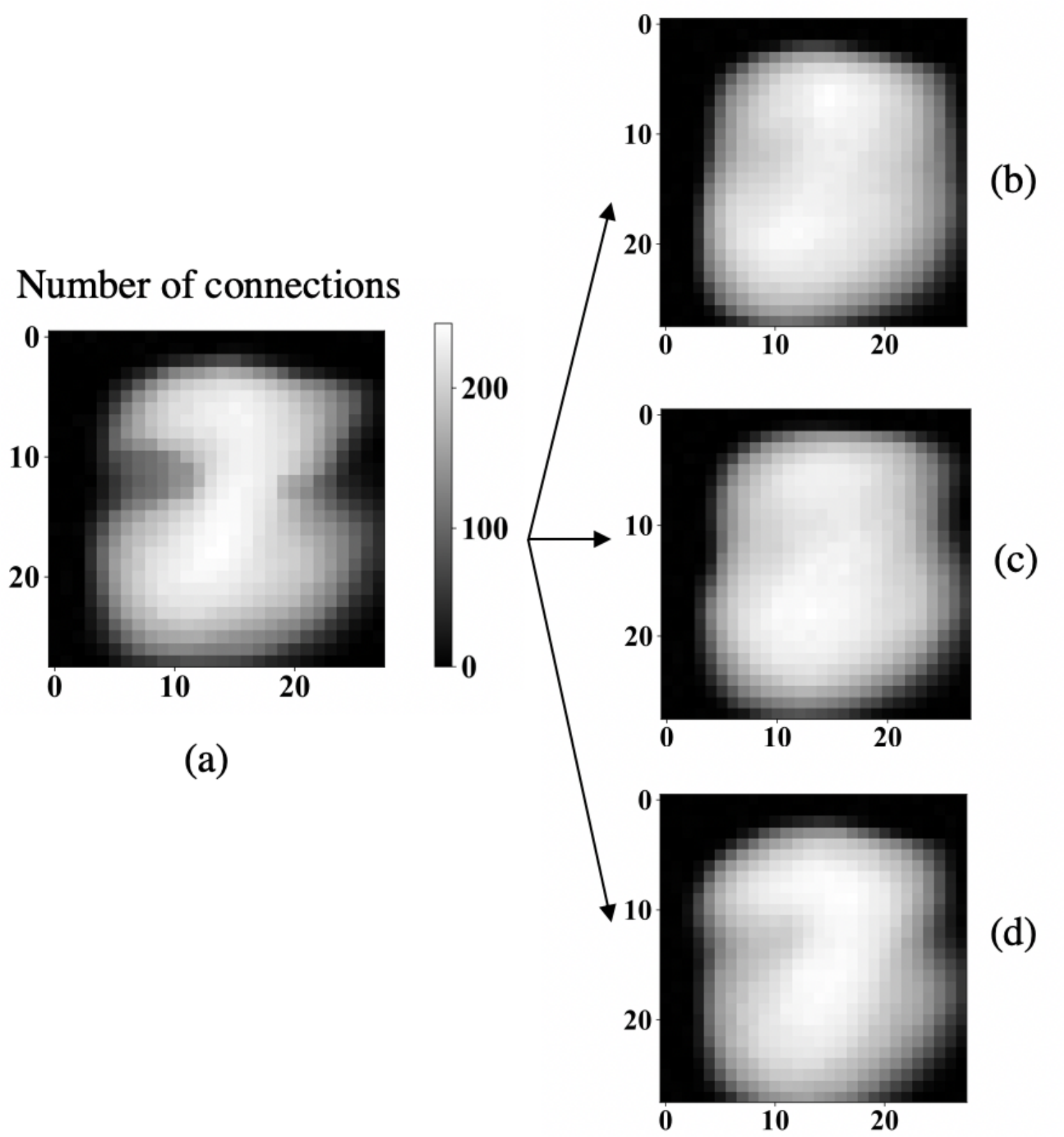}
\end{center}
\caption{Illustration of the grown connections from the input layer to the first hidden layer of 
LeNet-300-100 with different training data added.}
\label{fig:growth_phase}
\end{figure}

The connection growth policy effectively adapts the model architecture to accommodate newly available 
data and information.  To illustrate this, we extract and plot the total number of connections from 
each input image pixel to the first hidden layer of the post-growth LeNet-300-100~\cite{LeNet} (for 
the MNIST dataset, in which the images are hand-written digits of size 28$\times$ 28) in 
Fig.~\ref{fig:growth_phase}.  The initial model [Fig.~\ref{fig:growth_phase}(a)] is trained with 
data that has the label `1' or `2', and thus the connection density distribution is similar to an 
overlap of digits `1' and `2'.  Then, we add an additional class with labels `0', `6', and `7', and 
plot the corresponding connection density distribution of the post-growth network in 
Fig.~\ref{fig:growth_phase}(b), (c), and (d), respectively.  We observe that the network architecture 
evolves to adapt to the new class of data.

\subsubsection{Growth on new data}
To reduce the training cost of a model update, we introduce a mechanism to speed up the growth 
phase.  Specifically, we first employ connection growth and parameter training only on the previously 
unseen data for a pre-defined number of epochs whenever new data become available.  Then, we merge 
the new data with all the previously available training data, and perform growth and training on all 
existing data.

This `new data first' policy enables a rapid learning process and architecture update on the new data 
and significantly reduces overall training cost in the growth phase.  We compare the number of 
training epochs for LeNet-300-100 in Table~\ref{tab:epoch_comp} using two different approaches:
\begin{itemize}
\item \textbf{Merged training}: Merge the new data and existing data, and conduct connection growth 
and parameter training on all data.
\item \textbf{New data first}: Perform growth and training on new data first, then combine the new 
data and existing data, and finally grow and train on all available data.
\end{itemize}
In Table~\ref{tab:epoch_comp}, the initial model is trained on 90\% of the MNIST training data. New 
data and all data refer to the remaining 10\% of training data and the entire MNIST training set, 
respectively. To reach the same target accuracy of 98.67\%, our proposed method only requires 15 and 
20 training epochs first on new data and then on all data, respectively.  Since the number of training 
instances in new data is 10$\times$ smaller than in all data, the cost of 15 training epochs on new 
data is equivalent to only 1.5 epochs of training on all data. Thus, the training cost of our proposed approach (15 epochs on new data plus 20 epochs on all data) is equivalent to 21.5 epochs of training on all data.  As a result, our proposed method 
reduces the growth phase training cost by 2.3$\times$ compared to merged training, which requires 49 epochs of training on all data.

\begin{table}[t]
\caption{Training cost comparison between merged training and `new data first' approaches}
\label{tab:epoch_comp}
\begin{center}
\begin{tabular}{lcccc}
\hline
Approach & \multicolumn{2}{c}{\#Training epochs} & Accuracy \\
& New data$^{1}$ & All data & \\
\hline
Merged training & - & 49 & 98.67\% \\
New data first (ours) & 15 & 20 & 98.67\% \\
\hline
\multicolumn{4}{l}{\scriptsize{1: The size of new data is 10$\times$ smaller than the size of all data.}}\\
\end{tabular}
\end{center}
\end{table}

\subsection{Pruning phase}
We discuss the pruning approach next.

\subsubsection{Magnitude-based pruning policy}
DNNs are typically very over-parameterized.  Pruning has been shown to be very effective in
removing redundancy~\cite{PruningHS}.  Thus, we prune away redundant connections for compactness and 
to ensure efficient inference after the growth phase.

The pruning policy removes weights based on their magnitudes.  In the pruning process, we remove a 
connection $w$ by setting its value as well as the value of its corresponding mask to 0 if and 
only if the following condition is satisfied:
\begin{equation}
|w| \leq \beta ^ {th} \text{percentile of} \ \ |\textbf{W}|
\end{equation}
where $\beta$ is a pre-defined pruning ratio.  Typically, we use $3\leq \beta \leq 5$ in our 
experiments.  Note that connection pruning is an iterative process.  In each iteration, we prune 
the weights that have the smallest values (e.g., smallest 5\%), and retrain the network to recover 
its accuracy.  Once the desired accuracy is achieved, we start the next pruning iteration.

\subsubsection{Recoverable and non-recoverable pruning}
It is important for the incremental learning framework to be sustainable and support long-term 
learning. This is because we need to update the model frequently for a long period of time in many 
real-world scenarios. In such settings, the growth and pruning process needs to be executed 
over numerous cycles.  To support long-term learning, the gradient-based growth phase should be 
able to fully recover the network capacity, architecture, and accuracy from the last post-pruning 
model.  To achieve this, we employ recoverable pruning in the main grow-and-prune based model 
update process.  We explain the recoverable pruning policy next. 

We define a pruning process to be recoverable if and only if both of the following conditions are 
satisfied:
\begin{itemize}
\item \textbf{No neuron pruning}: Each neuron in the post-pruning network has at least one input 
connection and one output connection.  This ensures gradient flow in the growth phase in the next 
update.
\item \textbf{No accuracy loss}: The post-pruning model has the same or higher accuracy than the 
pre-pruning model.  This prevents information loss in the pruning phase.
\end{itemize}

In addition, we use a leaky rectified linear unit (ReLU) with a reverse slope of 0.01 as the 
activation function $f$ in the entire model update process:
\begin{equation}
f(x) = max(0.01x, x)
\end{equation}
This prevents the `dying' neuron problem (a ReLU with constant 0 output has no back-propagated 
gradient).  It keeps all the neurons active and thus the number of neurons does not decrease even 
after numerous cycles of growth and pruning. 

Some real-world scenarios (e.g., real-time video processing on mobile platforms and local inference 
on edge devices) may have very stringent computation cost constraints~\cite{him}.  Thus, we 
introduce non-recoverable pruning as an optional post-processing step to trade in accuracy and 
recoverability for extreme compactness.  In this process, both conditions for recoverable pruning 
can be violated, and there is no guarantee that another gradient-based growth phase can fully recover 
the architecture.  However, non-recoverable pruning effectively shrinks the model size further with 
only a minor loss in accuracy in our experiments.  For example, it provides an additional 
1.8$\times$ compression on top of recoverable pruning on LeNet-300-100, with only a 0.07\% absolute 
accuracy loss on the MNIST dataset.  We provide a detailed comparison between the models derived from 
recoverable and non-recoverable pruning in Table~\ref{tab:recoverable}, where the error rates for 
ResNet-18 and DeepSpeech2 refer to the top-5 error rate and the word error rate (WER), respectively.

\begin{table}[t]
\caption{Comparison of models derived from recoverable and non-recoverable pruning}
\label{tab:recoverable}
\begin{center}
\begin{tabular}{lccccc}
\hline
Network & Pruning method & Error rate (\%) & \#Parameters \\
\hline
\multirow{ 2}{*}{LeNet-300-100} & Recoverable & 1.33 & 21.7K \\
 & Non-recoverable & 1.40 & 12.2K \\
 \hline
\multirow{ 2}{*}{LeNet-5} & Recoverable & 0.83 & 7.9K \\
 & Non-recoverable & 0.88 & 5.4K \\
 \hline
 \multirow{ 2}{*}{ResNet-18}  & Recoverable & 11.12 & 3.9M \\
 & Non-recoverable & 11.25 & 2.8M \\
 \hline
 \multirow{ 2}{*}{DeepSpeech2} & Recoverable & 11.7 & 5.7M \\
 & Non-recoverable & 12.4 & 3.1M \\
\hline
\end{tabular}
\end{center}
\vskip -0.1in
\end{table}

\section{Experimental results}\label{sec:experiments}
We implement our framework using PyTorch~\cite{pytorch} on Nvidia GeForce GTX 1060 GPU (with 
1.708 GHz frequency and 6 GB memory) and Tesla P100 GPU (with 1.329 GHz frequency and 16 GB memory).  
We employ CUDA 8.0 and CUDNN 5.1 libraries in our experiments.  We report our experimental results 
for image classification on the MNIST and ImageNet datasets as well as speech recognition on the 
AN4 dataset.  

\begin{figure}
\begin {center}
\includegraphics[width=80mm]{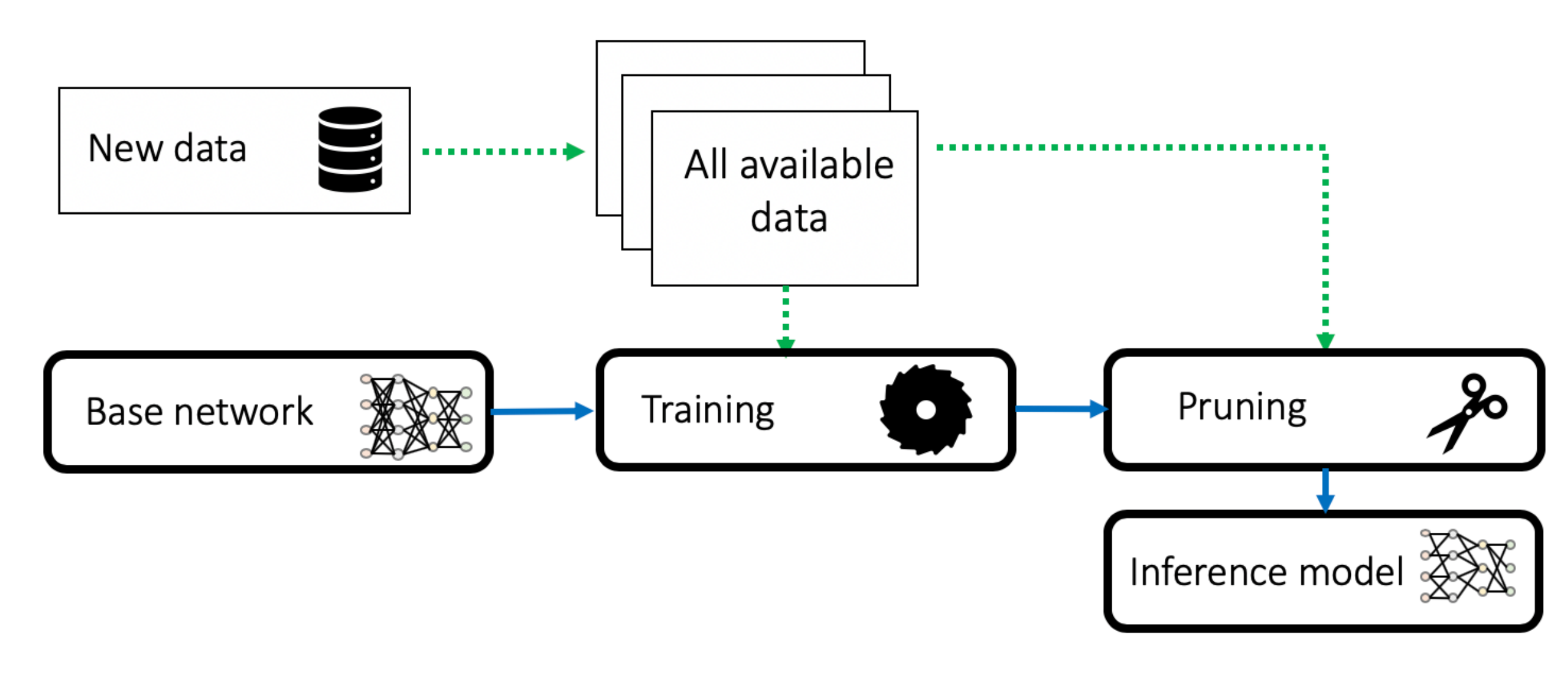}
\end{center}
\caption{Illustration of the training-from-scratch approach.}
\label{fig:train_from_scratch}
\end{figure}

\begin{figure}
\begin {center}
\includegraphics[width=80mm]{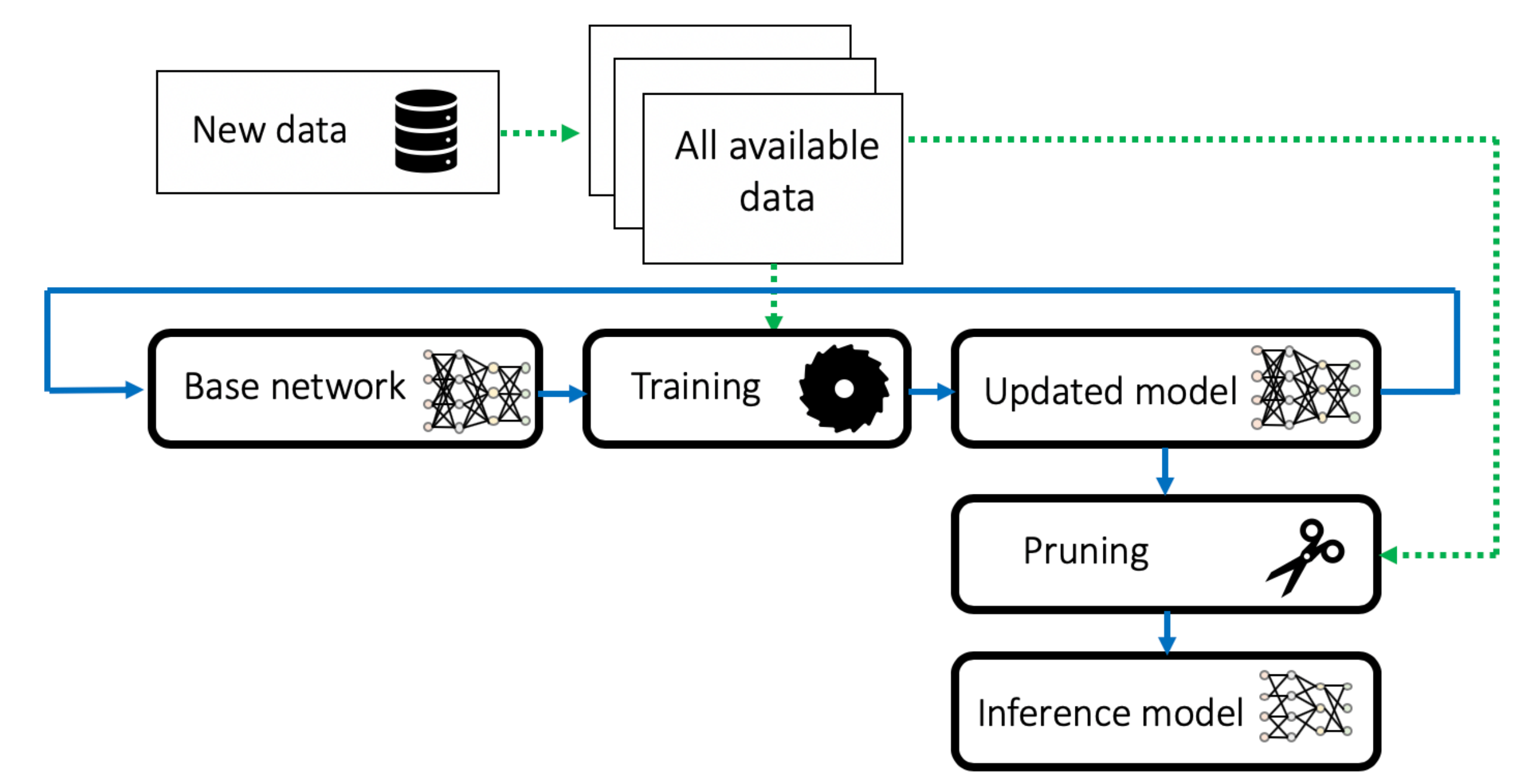}
\end{center}
\caption{Illustration of the network fine-tunning approach.}
\label{fig:network_finetunning}
\end{figure}

To validate the effectiveness of our method, we compare our proposed incremental learning framework 
with two widely-used conventional methods (TFS and NFT) \cite{incremental0, transferlearning1}.  We 
explain these two approaches next:
\begin{itemize}
\item \textbf{TFS}: Whenever a model update is needed, we train a model from scratch with all 
available data, and then prune it for compactness.  We illustrate the TFS approach in 
Fig.~\ref{fig:train_from_scratch}.
\item \textbf{NFT}: We maintain a model with all the connections activated and train it on all 
available data whenever an update is required.  The generated model can be used for the next update.  
Then, we make a copy of the model and prune it for compactness.  We illustrate the NFT approach in 
Fig.~\ref{fig:network_finetunning}.
\end{itemize}

We provide details of the experimental settings and results next.

\subsection{LeNets on MNIST}
We first show the effectiveness of our proposed methodology using LeNet-300-100 and LeNet-5 on the 
MNIST dataset.

\begin{table*}[h!]
\centering 
\caption{Experimental results for LeNet-300-100 on the MNIST dataset}
\begin{tabular}{l|p{0.8cm}<{\centering}p{0.8cm}<{\centering}p{0.8cm}<{\centering}|p{0.8cm}<{\centering}p{0.8cm}<{\centering} p{0.8cm}<{\centering}|p{0.8cm}<{\centering}p{0.8cm}<{\centering}p{0.8cm}<{\centering}rrrrr}
\hline
Training & \multicolumn{3}{c|}{Error rate (\%)}  & \multicolumn{3}{c|}{\#Parameters (K)} & \multicolumn{3}{c}{\#Training epochs} \\
data used & TFS & NFT & Ours & TFS & NFT & Ours & TFS & NFT & Ours \\
\hline
10\% & 2.27 & 2.27 & \textbf{2.19} & 19.4 & 19.4 & \textbf{13.9} & \textbf{141} & \textbf{141} & 166  \\
20\% & 1.80 & 1.84 & \textbf{1.70} & 22.7 & 20.5 & \textbf{17.6} & 131 & 136 &  \textbf{109} \\
30\% & 1.66 & 1.68 & \textbf{1.60} &  23.9 & 21.5 & \textbf{17.0}& 136 & 130&  \textbf{93} \\
40\% & 1.63 & 1.62 & \textbf{1.55}   & 20.5 & 22.7 & \textbf{13.8}& 131 & 124 & \textbf{79} \\
50\% & 1.58 & 1.66 &  \textbf{1.51}  & 26.5 & 23.9 & \textbf{17.0}&135 & 135 & \textbf{85}\\
60\% & 1.43 & 1.51 & \textbf{1.39}  & 27.8 & 29.3 & \textbf{26.5} & 140 & 109 & \textbf{71} \\
70\% & 1.41 & 1.42 & \textbf{1.35}  & 25.1 & 26.5 & \textbf{21.2}&139 & 133 & \textbf{70}   \\
80\% & 1.41 & 1.42 & \textbf{1.36}  & 29.3 & 27.8 & \textbf{21.1}& 141 &127 & \textbf{66}   \\
90\% & 1.41 & 1.43 & \textbf{1.33}  & 27.8 & 30.9 & \textbf{20.8} & 134 & 124 & \textbf{58} \\
100\%& 1.40 & 1.41 & \textbf{1.33}  & 31.3 & 32.5 & \textbf{21.7}& 137 & 131 & \textbf{49}  \\
\hline
\end{tabular}%
\label{tab:lenet3_res}
\end{table*}

\begin{table*}[h!]
\centering 
\caption{Experimental results for LeNet-5 on the MNIST dataset}
\begin{tabular}{l|p{0.8cm}<{\centering}p{0.8cm}<{\centering}p{0.8cm}<{\centering}|p{0.8cm}<{\centering}p{0.8cm}<{\centering} p{0.8cm}<{\centering}|p{0.8cm}<{\centering}p{0.8cm}<{\centering}p{0.8cm}<{\centering}rrrrr}
\hline
Training & \multicolumn{3}{c|}{Error rate (\%)}  & \multicolumn{3}{c|}{\#Parameters (K)} & \multicolumn{3}{c}{\#Training epochs} \\
data used & TFS & NFT & Ours & TFS & NFT & Ours & TFS & NFT & Ours \\
\hline
10\% & 1.62 & 1.62 & \textbf{1.45} & 5.3 & 5.3 & \textbf{4.9} & \textbf{148} & \textbf{148} & 162 \\
20\% & 1.28 & 1.30 & \textbf{1.23} & 5.9 & 6.5 & \textbf{5.5}& 141 & 126 & \textbf{110} \\
30\% & 1.01 & 1.05 & \textbf{0.97}& 6.5 & \textbf{6.2} & \textbf{6.2} & 136 & 118 & \textbf{92} \\
40\% & 0.97 & 0.99 & \textbf{0.92}  & 6.5 & 6.8 & \textbf{5.9}& 151 &  134 &  \textbf{80}  \\
50\% & 0.95 & 1.00 & \textbf{0.93} & 7.2 & 6.8 & \textbf{6.0} &135 & 141 &  \textbf{83}\\
60\% & 0.93 & 0.97 & \textbf{0.90}  & 8.8 & 7.6 & \textbf{7.5} & 120 & 119 &  \textbf{69} \\
70\% & 0.90 & 0.94 & \textbf{0.87}  & 8.0 & 8.4 & \textbf{6.3} & 129 & 99 &  \textbf{66}  \\
80\% & 0.90 & 0.90 & \textbf{0.87}   & 7.6 & 8.8 & \textbf{7.1} & 131 & 119 &  \textbf{52} \\
90\% & 0.88 & 0.91 & \textbf{0.87}  & 10.9 & 9.8 & \textbf{7.9} & 124 & 127 &  \textbf{56} \\
100\%& 0.88 & 0.90 &\textbf{0.83} & 10.3 & 10.2 & \textbf{7.9}& 129 & 115 &  \textbf{43}  \\
\hline
\multicolumn{10}{l}{\scriptsize{}}
\end{tabular}%
\label{tab:lenet5_res}
\end{table*}

\noindent
\textbf{Architectures}: We target two different base networks in the experiments: LeNet-300-100 and 
LeNet-5.  These two networks were proposed in~\cite{LeNet}.  LeNet-300-100 is an MLP neural network.  
It has two hidden layers with 300 and 100 neurons each.  LeNet-5 is a CNN with four hidden layers 
[two convolutional and two fully-connected (FC) layers].  The two convolutional layers share the same 
kernel size of 5$\times$5 and contain 6 and 16 filters, respectively, whereas the two FC layers have 
120 and 84 neurons, respectively.  The total number of network parameters in LeNet-300-100 and 
LeNet-5 is 266K and 59K, respectively.

\noindent
\textbf{Dataset}: We report results on the MNIST dataset~\cite{LeNet}.  It has 70K (60K for training 
and 10K for testing) hand-written digit images of size 28$\times$28.  We randomly reserve 5K images 
from the training set to build the validation set.  We introduce affine distortions to the 
training instances for data augmentation, same as in~\cite{LeNet}.

\noindent
\textbf{Training}: We split the training set (with 55K images) randomly into ten different parts 
of equal size.  In the incremental learning experiments, we start with one part to train the 
initial model for subsequent updates. We then add one part as new data each time in the incremental 
learning scenario.  For each update, we perform growth on new data and all data for 15 epochs and 
20 epochs in the growth phase, respectively.  Then, we prune the post-growth network for compactness.  
As for the two baselines (TFS and NFT), we train the model for 60 epochs, then prune the model 
iteratively.  Note that all the models share the same recoverable pruning policy for a fair 
comparison of model size.

We compare the test error rate, number of parameters, and number of training epochs for the three 
approaches on LeNet-300-100 and LeNet-5 in Table~\ref{tab:lenet3_res} and Table~\ref{tab:lenet5_res}, 
respectively.  We execute incremental learning ten times in our experiments.  In all the cases 
except the first round, our proposed method simultaneously delivers higher accuracy, reduced or 
equal model size, and less training cost relative to both baseline approaches.  For example, when we 
add the last 10\% training data to the existing 90\% data, our grow-and-prune paradigm based 
incremental learning algorithm leads to 0.07\% (0.08\%) absolute accuracy gain, 31\% (33\%) model 
size reduction, and 64\% (63\%) training cost reduction compared to the conventional TFS (NFT) 
approach on LeNet-300-100.  We observe similar improvements on LeNet-5.

Note that our incremental learning framework has higher training cost for the initial model (where 
only 10\% training data are available).  This is as expected since there is no existing model or 
knowledge for the initial model to start from, and thus all three approaches have to employ random 
initialization and start from scratch (TFS is equivalent to NFT in this case).  However, whenever a 
pre-trained model with existing knowledge is available, our incremental learning approach always 
produces reduced training cost due to its capability of preserving existing knowledge effectively 
and distilling knowledge from new data efficiently.

\subsection{ResNet-18 on ImageNet}

\begin{table*}[h]
\centering 
\caption{Experimental results for ResNet-18 on the ImageNet dataset}
\begin{tabular}{l|p{0.8cm}<{\centering}p{0.8cm}<{\centering}p{0.8cm}<{\centering}|p{0.8cm}<{\centering}p{0.8cm}<{\centering} p{0.8cm}<{\centering}|p{0.8cm}<{\centering}p{0.8cm}<{\centering}p{0.8cm}<{\centering}rrrrr}
\hline
Training & \multicolumn{3}{c|}{Top-5 error rate (\%)}  & \multicolumn{3}{c|}{\#Parameters (M)} & \multicolumn{3}{c}{\#Training epochs} \\
data used & TFS & NFT & Ours & TFS & NFT & Ours & TFS & NFT & Ours \\
\hline
10\% & 30.40 & 30.40 & \textbf{30.36} & 2.8 & 2.8 & \textbf{1.9} & \textbf{229} & \textbf{229} & 269 \\
20\% & 21.48 & 21.52 & \textbf{21.37} & 2.9 & 3.0 & \textbf{2.1}& 239 & 232 &  \textbf{221} \\
30\% & 17.88 & 17.90 & \textbf{17.78}& 3.1 & 2.8 & \textbf{2.5} & 204 & 209 &  \textbf{170} \\
40\% & 15.50 & 15.49 & \textbf{15.44}  & 3.3 & 2.9 & \textbf{2.4}& 227 & 217 & \textbf{159}  \\
50\% & 14.09 & 14.17 & \textbf{14.03}& 3.5 & 3.5 & \textbf{2.6} & 260 & 240 & \textbf{130} \\
60\% & 12.77 & 12.90 & \textbf{12.70} & 3.5 & 3.5 & \textbf{2.9} & 231 & 195 & \textbf{139}  \\
70\% & 12.17 & 12.31 & \textbf{12.15} & 3.9 & 4.1 & \textbf{3.3} &198 & 206 & \textbf{109}   \\
80\% & 11.51 & 11.52 & \textbf{11.47} & 3.9 & 4.1 & \textbf{3.4} & 207 & 199 & \textbf{99}   \\
90\% & 11.47 & 11.64 & \textbf{11.44} & 4.3 & 4.9 & \textbf{3.3} & 259 & 187 & \textbf{92}  \\
100\%& 11.25 & 11.27 &\textbf{11.12} & 4.7 & 5.2 & \textbf{3.9}& 241 & 213 & \textbf{86}  \\
\hline
\multicolumn{10}{l}{\scriptsize{}}
\end{tabular}%
\label{tab:resnet_res}
\end{table*}

We now scale up the network architecture to ResNet-18 and the dataset to ImageNet, which is a 
widely-used benchmark for image recognition.

\noindent
\textbf{Architecture}: ResNet is a milestone CNN architecture~\cite{ResNet}.  The introduced residual 
connections alleviate the exploding and vanishing gradient problem in the training of DNNs with
large depth, and yield substantial accuracy improvements.  We use ResNet-18 as the base network in 
our experiment.  It has 17 convolutional layers and one FC layer.  The total number of parameters in 
ResNet-18 is 11.7M.

\noindent
\textbf{Dataset}:  We report the results on the ImageNet dataset~\cite{imageNetdataset}.  This is a 
large-scale dataset for image-classifying DNNs.  It has 1.2M and 50K images from 1,000 distinct 
categories for training and validation, respectively.  Since there is no publicly available test 
set, we randomly withhold 50 images from each class in the training set to build a validation set 
(50K images in all), and use the original validation set as the test set.  We report the test 
accuracy in our experiment.

\noindent
\textbf{Training}: Similar to the previous experiments on the MNIST dataset, we separate the training 
set evenly and randomly into ten different chunks.  We use one chunk as the initially available data 
and add one chunk as new data each time.   We perform growth on new data and all data for 20 epochs 
and 30 epochs in the growth phase in our proposed approach, respectively. We train the model for 
90 epochs for the two baselines.  In the pruning phase, all methods share the same recoverable 
pruning policy for a fair model size comparison.

Table~\ref{tab:resnet_res} compares three different metrics (top-5 error rate, number of parameters, 
and number of training epochs) for the three different approaches.  Our proposed approach again 
outperforms both baselines except the training cost for the initial model. For example, it achieves 
up to 64\% (60\%) training cost reduction for model update compared to the TFS (NFT) approach, while 
delivering 0.13\% (0.15\%) higher top-5 accuracy, and 17\% (25\%) smaller model size at the same time.

\subsection{DeepSpeech2 on AN4}

\begin{table*}[h!]
\centering 
\caption{Experimental results for DeepSpeech2 with H-LSTM on the AN4 dataset}
\begin{tabular}{l|p{0.8cm}<{\centering}p{0.8cm}<{\centering}p{0.8cm}<{\centering}|p{0.8cm}<{\centering}p{0.8cm}<{\centering} p{0.8cm}<{\centering}|p{0.8cm}<{\centering}p{0.8cm}<{\centering}p{0.8cm}<{\centering}rrrrr}
\hline
Training & \multicolumn{3}{c|}{WER (\%)}  & \multicolumn{3}{c|}{\#Parameters (M)} & \multicolumn{3}{c}{\#Training epochs} \\
data used & TFS & NFT & Ours & TFS & NFT & Ours & TFS & NFT & Ours \\
\hline
40\% & 48.1 & 48.1 & \textbf{45.8}  & 4.8 & 4.8 & \textbf{4.0}& \textbf{267} & \textbf{267} & 313  \\
50\% & 35.6 & 37.7 & \textbf{34.7}  & 4.7 & 5.1 & \textbf{4.0}& 251 & 241 & \textbf{189}\\
60\% & 33.9 & 34.0 & \textbf{32.4}  & 5.5 & 5.3 & \textbf{4.4}& 281 & 220 & \textbf{160}  \\
70\% & 25.1 & 23.5 & \textbf{22.9}  & 6.0 & 6.0 & \textbf{5.2} & 247 & 249 & \textbf{127}  \\
80\% & 20.9 & 21.8 & \textbf{20.1}  & 5.9 & 6.7 & \textbf{4.8}& 230 & 196 & \textbf{129}   \\
90\% & 15.8 & \textbf{15.3} & 15.4  & 7.0 & 8.2 & \textbf{5.7}& 272 & 219 & \textbf{101}  \\
100\%& 12.4 & 12.6 &\textbf{11.7}& 7.4 & 7.6 & \textbf{5.7}& 269 & 232 & \textbf{88}   \\
\hline
\multicolumn{10}{l}{\scriptsize{}}
\end{tabular}%
\label{tab:speech_res}
\end{table*}

We now consider another important machine learning application: speech recognition.  
We target the DeepSpeech2~\cite{deepspeech2} architecture with an H-LSTM on the 
AN4 dataset~\cite{an4}, and provide experimental details next.

\noindent
\textbf{Architecture}: DeepSpeech2 is a popular architecture for speech recognition.  
It has three convolutional layers, three recurrent layers, one FC layer, and one 
connectionist temporal classification layer~\cite{hlstm}.  The inputs of the network 
are Mel-frequency cepstral coefficients of the sound power spectrum. We use 
bidirectional H-LSTM recurrent layers in our experiments and set the hidden state 
width for the H-LSTM cells to 800, same as reported in~\cite{hlstm}.  We introduce 
a dropout ratio of 0.2 for the hidden layers in the H-LSTM cells.

\noindent
\textbf{Dataset}: The speech recognition dataset in our experiment is the AN4 
dataset~\cite{an4}, which has 948 and 130 utterances for training and validation, 
respectively.  We randomly reserve 100 utterances from the training set as the 
validation set, and use the original validation set as the test set. 

\noindent
\textbf{Training}: We first divide the training set evenly and randomly into ten different parts.  
We start with training an initial model based on partial training data, and then 
update the model based on the remaining parts. To train an initial model with 
acceptable accuracy, we find the minimum amount of training data to be 40\% of all 
available training data (i.e., four parts). A decrease in this amount leads to an 
abrupt drop in accuracy ($>$80\% WER when only three parts are used). Then, we add 
one part each time to update the model. For the model growth phase, we first grow 
the network for 20 epochs based on only the newly added data, and then 30 epochs 
when the new part is merged with existing ones.  We train the model for 120 epochs 
for both conventional baselines. We conduct recoverable pruning for all the methods 
in pursuit of model compactness.

We compare the WER and number of parameters for the models derived from the 
three different approaches as well as their corresponding training epochs in 
Table~\ref{tab:speech_res}.  We observe a significant improvement in the trade-offs 
among accuracy, model size, and training cost in our proposed incremental learning 
framework.  For example, when we add the last 10\% training data, our model achieves 
0.7\% (0.9\%) lower WER and 30\% (33\%) additional compression ratio with 
67\% (62\%) less training cost compared to the TFS (NFT) approach.

\section{Discussions}\label{sec:discussions}
In this section, we discuss the learning mechanism of human brains and the underlying 
inspirations of our proposed incremental learning framework.

Our brains are plastic. They continually remold numerous synaptic connections as 
we acquire new knowledge and information~\cite{neural}. Recent discoveries in 
neuroscience have unveiled the fact that our brain's synaptic connections change 
every second over our entire lifetime.   Furthermore, it has been shown that most 
new knowledge acquisition and information learning process in our brains result from 
the brain remodeling and synaptic connection rewiring mechanism (referred to as 
`neuroplasticity')~\cite{neural}. This is very different from the current DNNs
that have fixed architectures with weights trained only with back-propagation to 
distill intelligence from a dataset.

To mimic the learning mechanism of human brains, we utilize gradient-based growth 
and magnitude-based pruning to train DNN architectures in our framework.  During 
training, we adaptively adjust the connectivity of synaptic connections for the 
model to accommodate previously unseen data and acquire new knowledge.  Such 
a brain-inspired algorithm delivers an efficient and accurate inference model at
a significantly reduced learning cost.  Future work could entail dynamic rewiring 
to adapt DNN depth as well as architecture learning for arbitrary feedforward 
neural networks.

\section{Conclusions}\label{sec:conclusion}
In this paper, we proposed a brain-inspired incremental learning framework based on 
a grow-and-prune paradigm.  We combine gradient-based growth and magnitude-based 
pruning in the model update process.  We show the effectiveness and efficiency of 
our proposed methodology for different tasks on different datasets.  For 
LeNet-300-100 (LeNet-5) on the MNIST dataset, we cut down the training cost by up 
to 64\% (67\%) compared to the TFS approach and 63\% (63\%) compared to the NFT 
approach.  For ResNet-18 on the ImageNet dataset (DeepSpeech2 on the AN4 dataset), 
we reduce the training epochs by up to 64\% (67\%) compared to the TFS approach and 
60\% (62\%) compared to the NFT approach.  The derived models have improved 
accuracy (or reduced error rate) and more compact network architecture.

\ifCLASSOPTIONcaptionsoff
  \newpage
\fi

\bibliographystyle{IEEEtran} 
\bibliography{bibib}

\vfill
\clearpage
\end{document}